\documentclass{article}

     \usepackage[preprint]{neurips_2020}

\usepackage[utf8]{inputenc} 
\usepackage[T1]{fontenc}    
\usepackage{hyperref}       
\usepackage{url}            
\usepackage{booktabs}       
\usepackage{amsfonts}       
\usepackage{nicefrac}       
\usepackage{microtype}      

\usepackage{xspace}

\usepackage{overpic}
\usepackage{caption}        
\usepackage{subcaption}        
\usepackage{amsthm}
\usepackage{bbm}
\usepackage{amsmath}
\usepackage{mathtools}
\usepackage{amssymb}
\usepackage{float}
\usepackage{wrapfig}
\usepackage{comment}
\usepackage{pgfplots, pgfplotstable}
\usepackage{algorithm}

\usepackage{enumitem}
\usepackage{multirow}
\usepackage[title]{appendix}
\usepackage{listings}
\usepackage{xcolor}
\usepackage{wrapfig}

\definecolor{dkgreen}{rgb}{0,0.6,0}
\definecolor{gray}{rgb}{0.5,0.5,0.5}
\definecolor{mauve}{rgb}{0.58,0,0.82}

\definecolor{dark_green}{rgb}{0, 0.5, 0}

\definecolor{Fcolor}{HTML}{af2418}

\definecolor{Ccolor}{HTML}{ffd359}

\def\int{\mathrm{int}}

\renewcommand{\paragraph}[1]{\vspace{.1em}\noindent\textbf{#1.~}}

\usepackage{lipsum}

\title{Testing GLOM's ability to infer wholes from ambiguous parts}

\author{%
  Laura Culp \\
  Google AI\\
  \texttt{laculp@google.com} \\
  \And
  Sara Sabour\\
  Google AI\\
  \texttt{sasabour@google.com} \\
  \And
  Geoffrey E. Hinton\\
  Google AI\\
  \texttt{geoffhinton@google.com} \\
}

\begin{document}

\maketitle

\begin{abstract}
  The GLOM architecture proposed by \cite{hinton2021represent} is a recurrent neural network for parsing an image into a hierarchy of wholes and parts. When a part is ambiguous, GLOM assumes that the ambiguity can be resolved by allowing the part to make multi-modal predictions for the pose and identity of the whole to which it belongs and then using attention to similar predictions coming from other possibly ambiguous parts to settle on a common mode that is predicted by several different parts.  In this study, we describe a highly simplified version of GLOM that allows us to assess the effectiveness of this way of dealing with ambiguity. Our results show that, with supervised training, GLOM is able to successfully form islands of very similar embedding vectors for all of the locations occupied by the same object and it is also robust to strong noise injections in the input and to out-of-distribution input transformations. 
\end{abstract}
\section{Introduction}

The GLOM architecture is a recurrent neural network for vision introduced in \cite{hinton2021represent}. The neural hardware is divided into columns each of which is responsible for representing what is happening in a small patch of the image called a {\it location}. Within a column, there are multiple different embedding vectors that correspond to different levels of representation of what is occupying that location. For example, the embedding vectors in one column might represent a nostril, a nose, a face, a person, and a party. Each of these entities will typically occupy many locations and the central idea of GLOM is that all of the locations occupied by the same entity should use very similar embedding vectors for that entity. So all of the locations occupied by the face should have very similar embedding vectors {\it at that level of representation}, even though the locations occupied by the nose and the locations occupied by the mouth have very different representations at the next level down. The islands of similarity at the different levels of representation can be viewed as a parse tree that is dynamically created for each image without requiring any changes in the connectivity of the neural hardware. 

For each pair of adjacent levels within a column there is a local autoencoding loop that consists of a multi-layer, bottom-up neural net that acts as the encoder and a multi-layer top-down neural net that acts as the decoder, as shown in Figure~\ref{fig:ellipseglom_arch}. These neural nets are different for each pair of levels but are shared across all the columns. All of GLOM's knowledge resides in the shared neural networks that operate within a column.  Embeddings at the same level in different columns interact in a very simple way that does not have any learned parameters, though it may have adaptive hyper-parameters. The between-column interactions are simply attention-weighted averaging which makes the embedding at a particular level in one column regress towards the embeddings at the same level in other nearby columns but with more weight given to embeddings that are already similar. This can be viewed as a grossly simplified version of the transformer architecture \cite{Vaswani17} in which the embedding, query, key and value vectors are all identical.

\begin{figure}
    \centering
    \includegraphics[width=.7\textwidth]{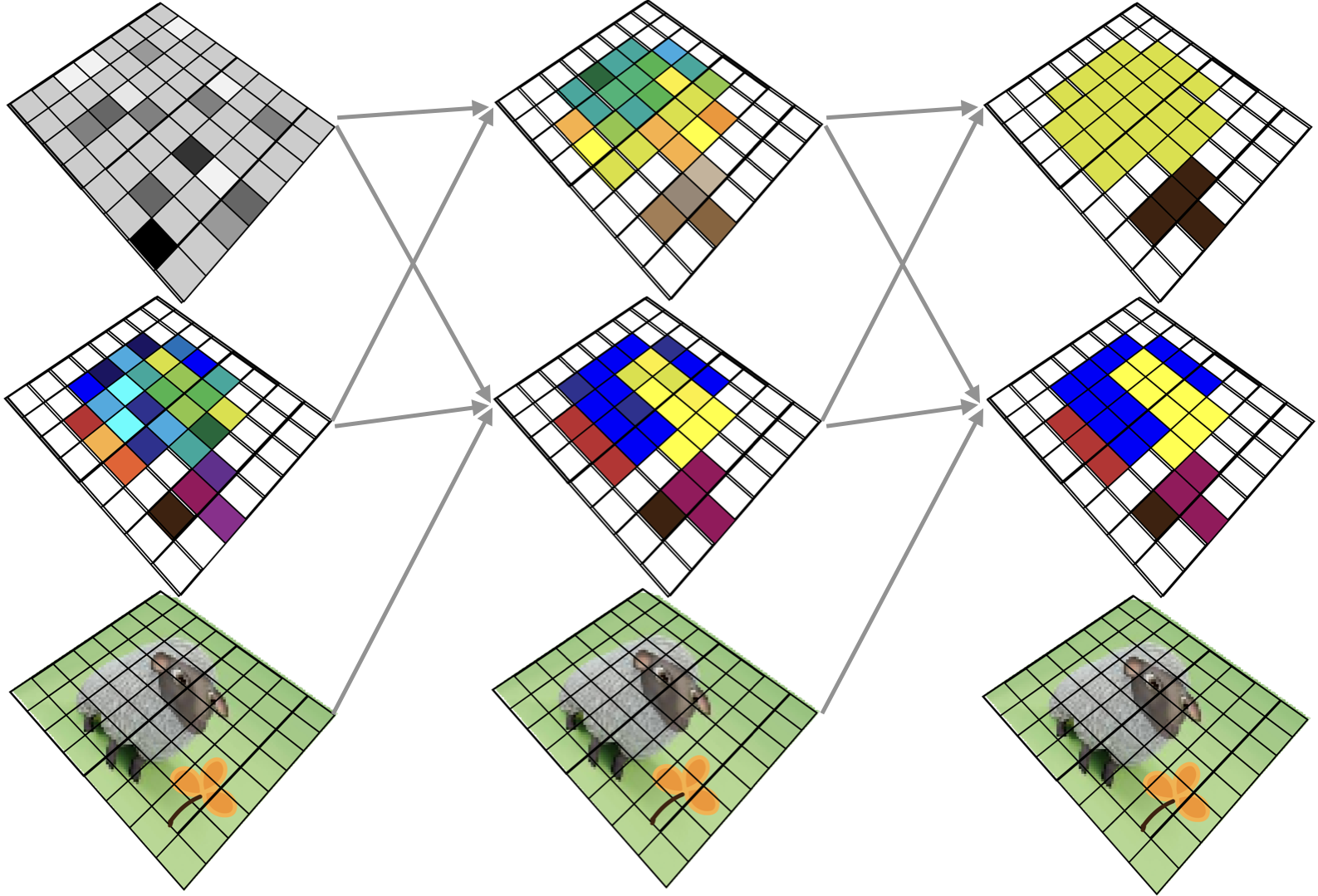}
    \caption{GLOM iteratively forms islands of similar embeddings at each level. At the final iteration the goal is to get near identical embeddings for all the columns that belong to the same object at the highest level.}
    \label{fig:islands}
\end{figure}

A static image is presented to GLOM as a sequence of identical frames and the columns of GLOM iteratively settle on embeddings at every level that are consistent with the embeddings at adjacent levels in the same column and also form spatially coherent islands of very similar embeddings at nearby locations, as shown in Figure \ref{fig:islands}.

\section{How GLOM deals with ambiguous parts}
\begin{wrapfigure}{r}{0.3\textwidth}
    \centering
    \includegraphics[width=.3\textwidth]{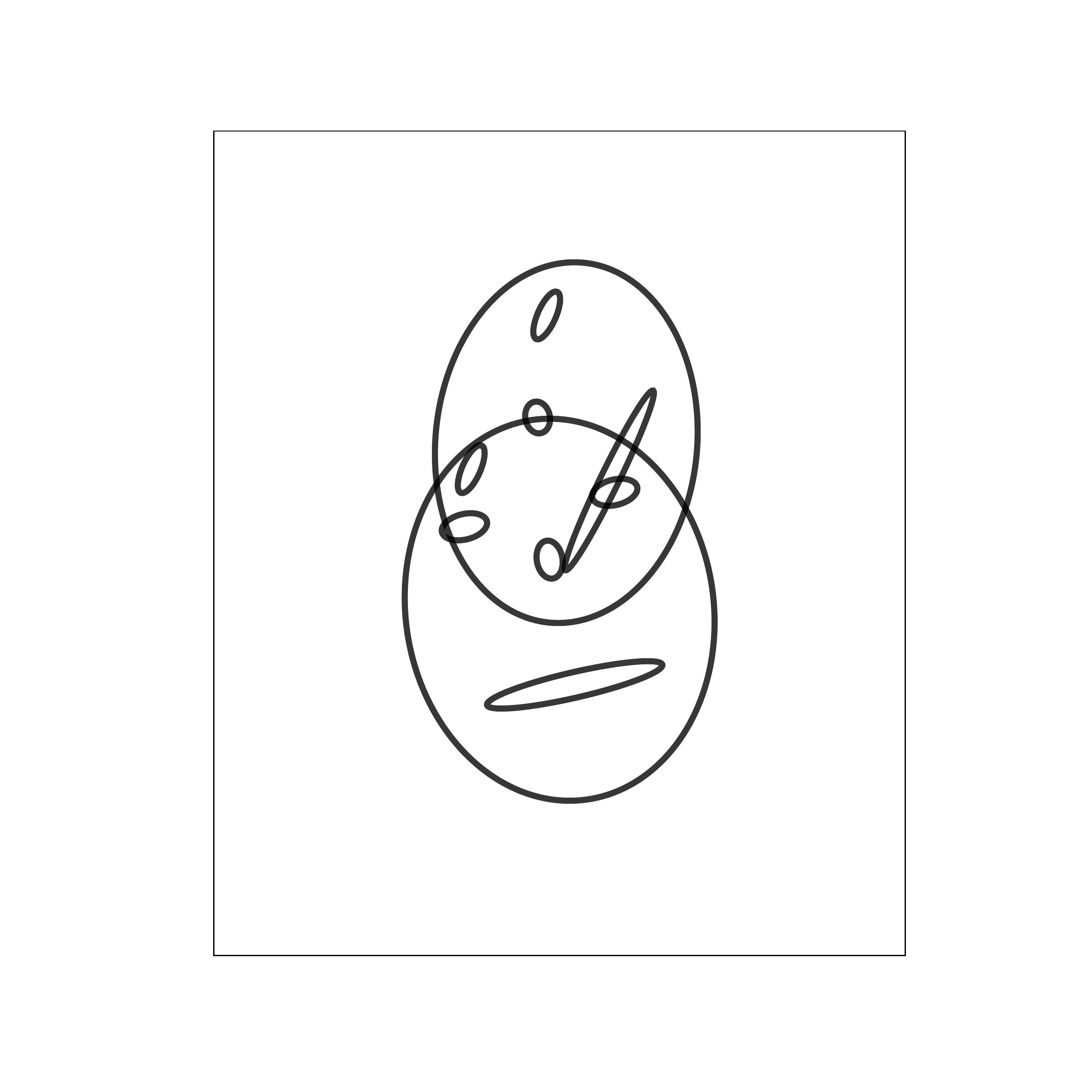}
    \caption{Two faces made up of five ellipses each. Together, the parts are highly ambiguous, but humans can use the spatial relationships between the parts to identify the whole.}
    \label{fig:sheepface}
\end{wrapfigure}

People can recognize shapes that are composed of highly ambiguous parts by using the spatial relationships between the parts, as shown in Figure \ref{fig:sheepface}. It seems that the deep neural nets that have been so successful for object classification rely less on these spatial relationships and more on high-frequency texture information thus making them susceptible to adversarial examples. One of the aims of GLOM is to design neural nets that are better at using the information in spatial relationships between parts. 

There are two rather different approaches to dealing with ambiguous parts. In the "random field" approach, the ambiguity is resolved by iterative interactions between the embedding vectors for the parts. This resembles the way that the meanings of word fragments are refined in BERT \citep{devlin2018bert}. At each layer, the embedding of a word fragment is refined by using the match between a query vector and a key vector to elicit relevant information from the current embeddings of other word fragments.  Unfortunately, this approach becomes much more complicated when dealing with the spatial relationships between parts because of the need to perform coordinate transforms:  For a potential nose to receive disambiguating support from a nearby potential mouth it is essential\footnote{Picasso demonstrated that it is not actually essential when the parts themselves are relatively unambiguous. Nevertheless, it is very helpful.} for the mouth to be in the right spatial relationship to the nose. So the nose would need to send out a query that specified the required pose of the mouth by applying the nose-to-mouth coordinate transform to the pose of the nose. Then the mouth would need to send back a value that applied the inverse coordinate transform to the pose of the mouth. 

An alternative way to disambiguate parts is to let each part make ambiguous, multi-modal predictions for the whole and then to look for a common mode in those predictions. This is a type of Hough transform \citep{hough1962method} and it replaces $O(N^2)$ non-local interactions involving coordinate transforms between parts with $O(N)$ local coordinate transforms between parts and wholes plus $O(N^2)$ very simple non-local interactions between the $N$ different multi-modal predictions for the wholes. This seems like a big win, but it relies on the ability of an ambiguous part to make an appropriate multi-modal prediction for the whole. The point of the experiments described in this paper is to test out whether GLOM can learn to make highly multi-modal predictions for the whole from highly ambiguous parts and whether simple attention-weighted averaging at the level of the whole can settle on the shared mode of these predictions.   

\section{Ellipse World}
To make it simpler to investigate the ability of GLOM to find the common mode in a set of learned multi-modal predictions, we have greatly simplified the problem of dealing with cluttered images. Instead of using pixel intensities as the raw input, we divide the image into a grid of locations and we assume that each location contains at most one ellipse center. For each location that contains an ellipse center, the input to our simplified version of GLOM, which we call eGLOM, is then a 6-dimensional vector that represents the ellipse. We call this vector the "ellipse symbol". The ellipse symbol contains the exact real-valued center coordinates, the rotation, and the scale of the ellipse. The scale of the ellipse can surpass the grid size and can be any real value. Locations that do not contain an ellipse center have no effect so the column of hardware that would normally be used to settle on a multilevel representation of whatever occupied that location does not need to be simulated \footnote{Assuming that we are not trying to fill in missing parts.}. 

We refer to the set of ellipses that form an image as the "image" using quotation marks to indicate that the image is not composed of pixels.

\begin{wrapfigure}{r}{0.5\textwidth}
    \vspace{-25pt}
    \centering
    \includegraphics[width=0.5\textwidth]{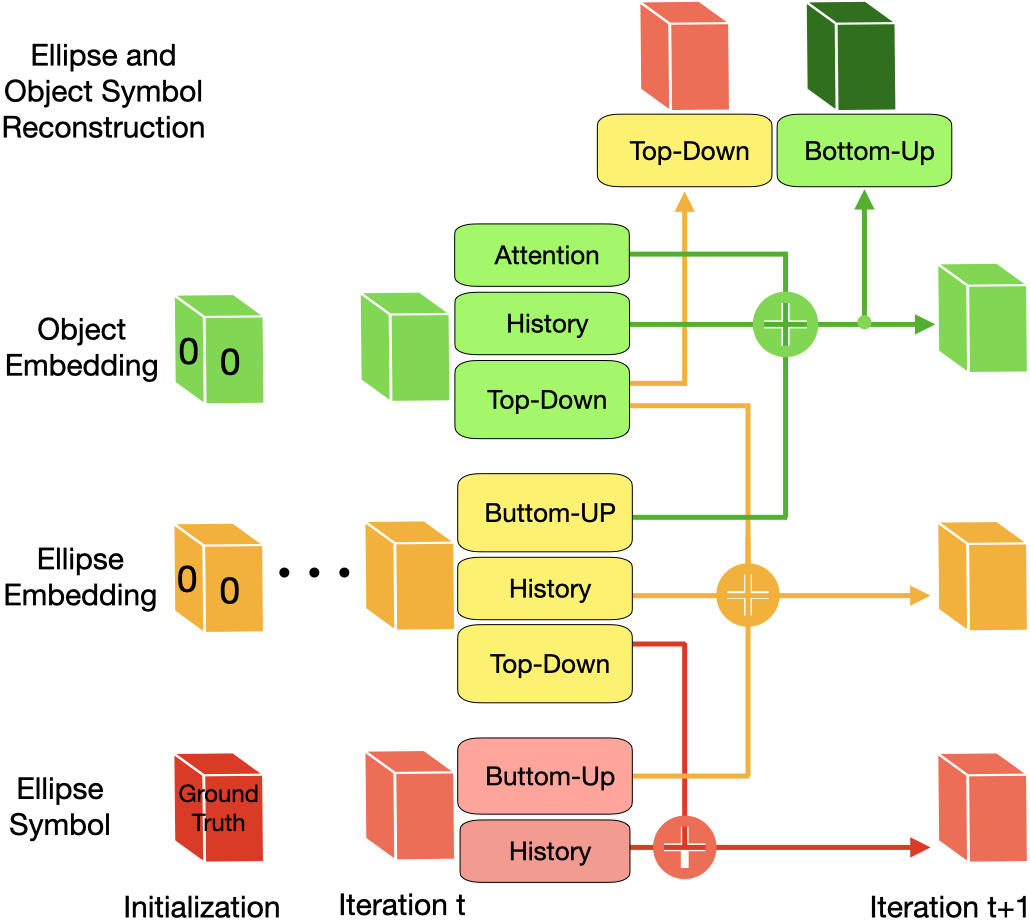}
    \caption{
    A single iteration of the eGLOM recurrent architecture. There are three levels of representation, containing bottom-up, top-down, attention, and history components as described in section 4.}
    \label{fig:eglom_arch}
    \vspace{-20pt}
\end{wrapfigure}
\section{Architecture}
 Two hierarchical levels of embedding are used at each occupied location in eGLOM. The lower level is for parts, all of which are ellipses, and the higher level is for whole objects all of which are composed of five ellipses. It is the interaction between these two levels of embedding that is main target of investigation.

 In addition, there are input and output representations which can be viewed as interfaces that are needed for the input and output of eGLOM, but do not correspond to anything in the full GLOM architecture. The input representations are the ellipse symbols described above, and the output representations are the object symbols that consist of 6 parameters for the pose of the object plus an N-way softmax for the identity of the object, where N is the number of types of object.  The object symbols are used for supervised training and for drawing eGLOM's interpretation of the "image".
 
 On each iteration in the recurrent architecture of eGLOM, the object embedding, ellipse embedding, and ellipse symbol are updated for each location. Figure~\ref{fig:eglom_arch} visualizes the recurrent steps for eGLOM. The ellipse embedding level is updated by combining bottom-up information from the ellipse symbol, top-down information from the object embedding, and the previous embedding at the ellipse level. The embedding at the object level is updated by combining bottom-up information from the ellipse embedding, the previous embedding at the object level and information coming from the object level embeddings at other locations via the attention mechanism. The same combined information is also used to predict the object symbol which allows supervised training. To allow us to visualize the object embedding, it generates the output ellipse symbol at its location by using the composition of the top-down networks from the object embedding to the ellipse embedding and from the ellipse embedding to the ellipse symbol. To allow for correction of perturbed ellipses, the input ellipse symbol at a location is updated by combining top-down information from the ellipse embedding at that location with the previous input ellipse symbol. Each of these components will be described in more detail below.
 
 In our experiments, the dimensionality of the ellipse and object embedding vectors is tuned as a hyper=parameter, and is at least 100.

\paragraph{Weight Sharing}
In eGLOM architecture, similar to transformers, all the operations and networks are replicated over all the grid positions, i.e. individual ellipses. Therefore, eGLOM has a permutation invariant architecture with MLPs as basic modules. One can view eGLOM operations also as 1x1 Convolution operations. The high degree of parameter sharing reduces the model complexity with the hope of improving the model's generalization.

\paragraph{The Bottom-up neural nets}
There are three bottom-up MLPs replicated with shared weights over grid cells. Each MLP has ReLU non-linearity and is applied at every grid location at each iteration.
The MLP from the Ellipse Symbol (Level 0) to the Ellipse Embedding (Level 1) has hidden layers of size 32 and 64. The MLP from the Ellipse Embedding (Level 1) to the Object Embedding (Level 2) has a single hidden layer of the same size as the embedding dimensions. The MLP from the Object Embedding (Level 2) to Object Symbol (level 3) has two hidden layers of size 64 and 32. 

\begin{wrapfigure}{r}{0.3\textwidth}
    \centering
    \includegraphics[width=0.3\textwidth]{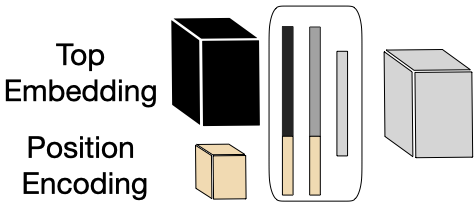}
    \caption{The Top-down MLPs concatenate a position encoding to the higher level representation and generate a lower level representation at that position.}
    \label{fig:td}
    \vspace{-10pt}
\end{wrapfigure}
\paragraph{The Top-down neural nets}
There are two top-down MLPs each with two hidden layers, Figure~\ref{fig:td}. The MLP from the Object Embedding to the Ellipse Embedding has hidden layers that are the same size as the embedding layers. The function of this MLP is to take an object-level embedding and a representation of the X-Y grid location and reconstruct an embedding of the part that is at that grid location. The MLP from the Ellipse Embedding to the Ellipse Symbol has hidden layers of decreasing size. Its function is to take an ellipse embedding and a representation of the location and reconstruct the ellipse symbol.

 As we are working within the continuous XY space for input and output (object and ellipse symbols), as opposed to pixel space ("images"), we grid the XY position to decide which location an ellipse belongs to. The grid XY position is assigned by rounding the coordinates on the X and Y dimensions. The pseudocode for this is `round(x or y / grid size) * grid size`.

Sine and cosine are used to transform the xy location into a continuous position encoding. We normalize the X and Y positions to a range of (-1, 1).  The sine and cosine of the normalized grid XY is appended to the object representation (or the embedding). In order to ensure the sine and cosine capture the highest frequency of detail, we also append the sine and cosine of 2* the normalized grid XY,  4*normalized XY, etc) before passing through a multi-layer neural net. 

\paragraph{Attention}
For updating the object embedding at each location based on the object embeddings at other locations at the previous time step, we use a grossly simplified version of the type of attention used in transformers. Instead of computing separate query, key, and value vectors from the current object embedding at a location, we simply compute the exponentiated scalar product of the object embedding with the object embeddings at all locations and then normalize to get attention weights that add to 1. The embeddings at all locations are then averaged using the attention weights and this average is combined with the bottom-up prediction to generate the output of that iteration for the object embedding at that location.  This eventually leads to the creation of islands of similarity, over multiple iterations. The scalar products used for computing the attention weights can be divided by a temperature and this hyper-parameter can be adjusted to increase or decrease the sharpness of the attention. 

\paragraph{Weighted Averaging}
During the iterations of the eGlom as a recurrent net the weights in the top-down and bottom-up neural nets do not change, but as the iterations proceed we change the relative balance of the top-down and bottom-up inputs used to compute the next ellipse level embeddings. We use a weighted average of these two inputs weighting one as more important than the other based on which we have more confidence in. At the first iteration, the object level embeddings are all initialized at zero so we combine the top-down and bottom-up by only taking the bottom-up input. By the second iteration, the object-level embeddings contain useful information, so the top-down inputs get some weight but most of the weight is still given to the bottom-up input. By the middle iteration, half the weight comes from the bottom-up and half from the top-down. On the final iteration, all the weight comes from the top-down, and none comes from bottom-up. This allows us to correct perturbed input ellipses.

For the object-level embeddings, there is no top down input. Instead, there is an input coming from all the other locations via the attention weights. The weight of the total attentional contribution to a location relative to the bottom-up and history contributions does not change over the iterations. 

In addition to just the bottom-up and top-down, we also consider the embedding at the same level on the previous iteration as \textbf{`history'}. This is meant to stabilize the network. 

In summary, the weight attributed to the history, bottom-up, and top-down for the ellipse level embedding all sum to one, with the history weight not changing over iterations. The history, attention, and bottom-up weights all sum to one for the object-level embedding, and do not change over iterations.

\begin{figure}[t]
    \centering
    \begin{subfigure}[t]{0.49\textwidth}
    \centering
    \includegraphics[width=.9\textwidth]{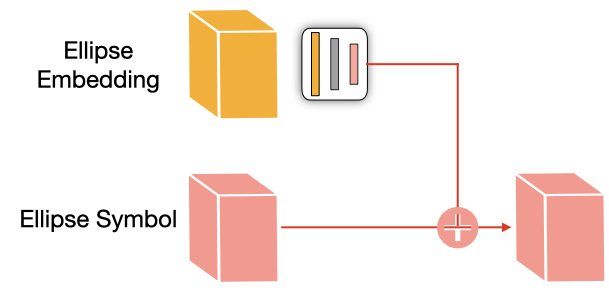}
    \caption{Ellipse Symbol}
    \label{fig:ellipse_symbol}
    \end{subfigure}
    \begin{subfigure}[t]{0.49\textwidth}
    \centering
    \includegraphics[width=.9\textwidth]{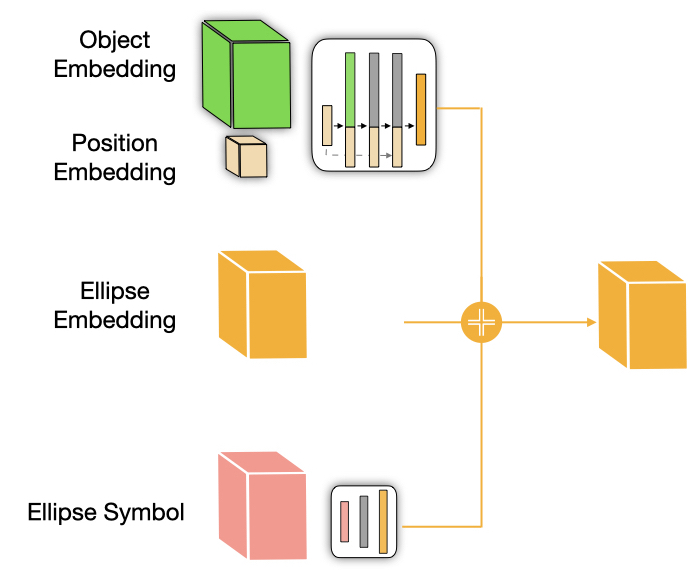}
    \caption{Ellipse Embedding}
    \label{fig:ellipse_embedding}
    \end{subfigure}
    \newline
    \newline
    \begin{subfigure}[t]{0.49\textwidth}
    \centering
    \includegraphics[width=.9\textwidth]{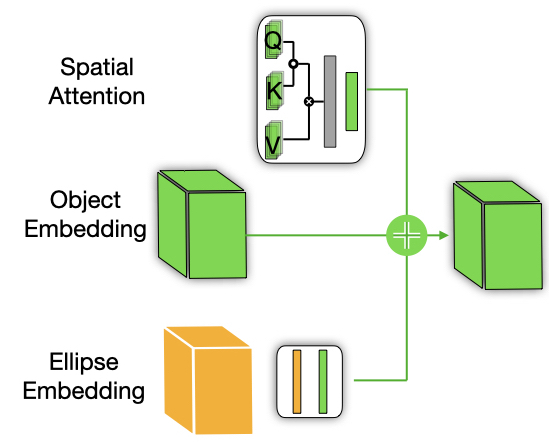}
    \caption{Object Embedding}
    \label{fig:object_embedding}
    \end{subfigure}
    \begin{subfigure}[t]{0.49\textwidth}
    \centering
    \includegraphics[width=.9\textwidth]{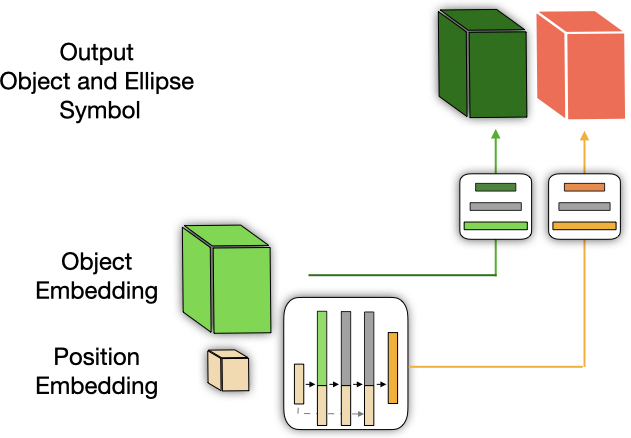}
    \caption{Reconstructed Symbols}
    \label{fig:object_symbol}
    \end{subfigure}
    \caption{Each subfigure visualizes connections at different levels of the hierarchy for calculating {\bf a}) ellipse symbols, {\bf b}) ellipse embeddings, {\bf c}) object embeddings, and {\bf d}) output reconstructed symbols.}
    \label{fig:glom_connections}
\end{figure}

\begin{figure}
    \begin{center}
    \includegraphics[width=\textwidth]{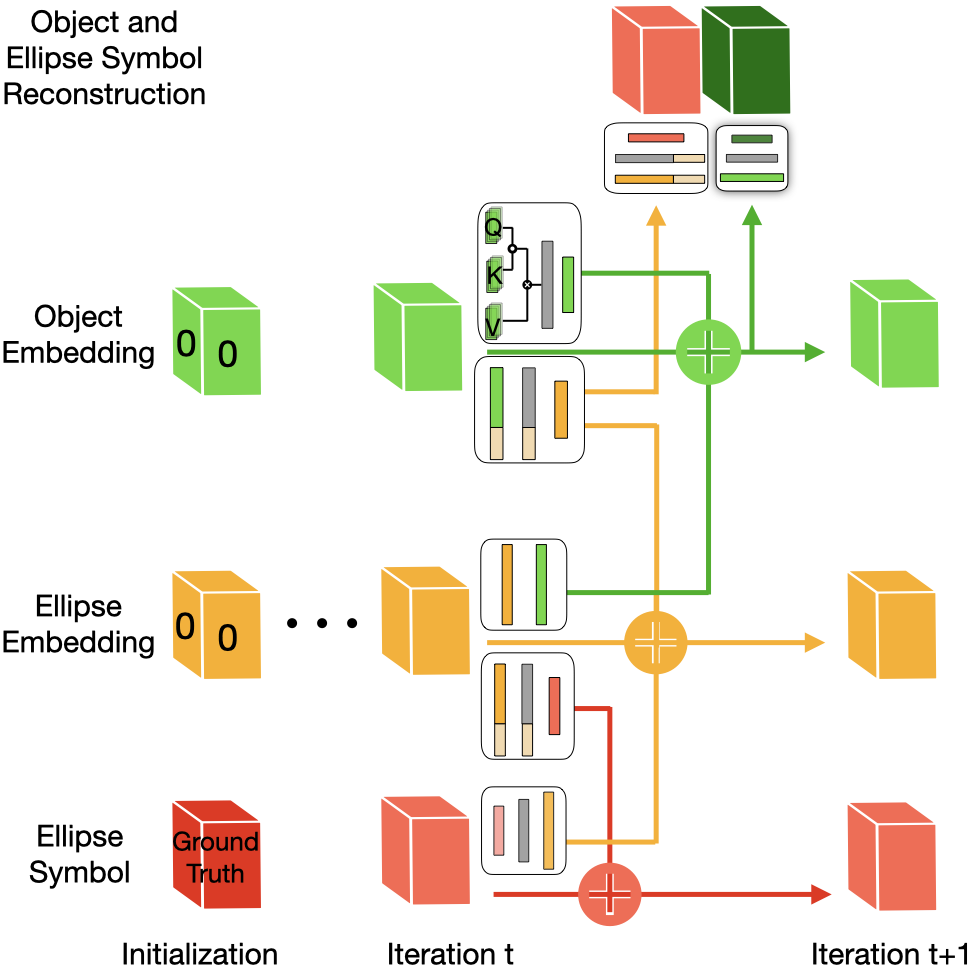}
    \caption{Glom has a recurrent architecture, where each cell at each level updates the same cell position at the lower levels, upper levels, and the local neighbours of it at the same level. The connection details are depicted in Figure ~\ref{fig:glom_connections}. At the first iteration embeddings are initialized to zero. At the last iteration, an entire top to bottom decoding is completed, to ensure that a full reconstruction can occur for the ellipse embeddings and symbols.}
    \label{fig:ellipseglom_arch}
    \vspace{-15pt}
    \end{center}
\end{figure}

\paragraph{Loss functions and Metrics}
The loss comes into play in two different places. There is a Reconstruction Loss based on the discrepancy between the initial ellipse symbols that are the raw input to eGLOM and the reconstructions of these symbols over time from the ellipse embeddings. 
The Reconstruction Loss is important to prevent drift, or, if perturbed ellipses are given, to allow for successful correction of the perturbed parameters of the ellipse symbol. It is calculated by first using the top-down net to compute an ellipse embedding from the object embedding and then using the lower-level top-down net to compute an ellipse symbol from the ellipse embedding. The computed ellipse symbol is then compared to the ground truth Ellipse Symbol, Figure~\ref{fig:object_symbol}. We use the mean squared error of the difference and, during training, the Reconstruction Loss is applied to the output of every iteration and backpropagated through time.

There is also a top-level Object Loss which has two parts. The first is a mean squared error between the 6 pose parameters of the true object symbol and the prediction of the bottom-up net operating on the object embedding. The second is the cross entropy loss between the true class of the object at that location and the probability distribution over the possible classes computed by the bottom-up net operating on the object-level embedding. The top-level Object loss is only applied at the last iteration, but it is back-propagated through time.

We report the metrics of the Mean Squared Error of the Object Symbols (the Whole MSE), the Classification Accuracy of the Object Types, and the Mean Squared Error of the Ellipse Symbols (the Part MSE) at the last iteration.

\paragraph{Regularizers}
In order to achieve better creation of islands, we add a regularizer that encourages the attention prediction coming from other locations and the bottom-up prediction coming from the ellipse embedding at that location to be more similar. After combining these two predictions to get the next object embedding, we minimize the cosine distance between the bottom-up prediction and the resulting object embedding.  Minimizing this cosine distance encourages the bottom-up neural net to make more similar predictions for the object embeddings at different locations if these embeddings are already sufficiently similar to dominate the attention weights. 

\subsection{Dataset}

\begin{figure}
    \centering
    \begin{subfigure}[b]{0.5\textwidth}
    \centering
    \includegraphics[width=0.9\textwidth]{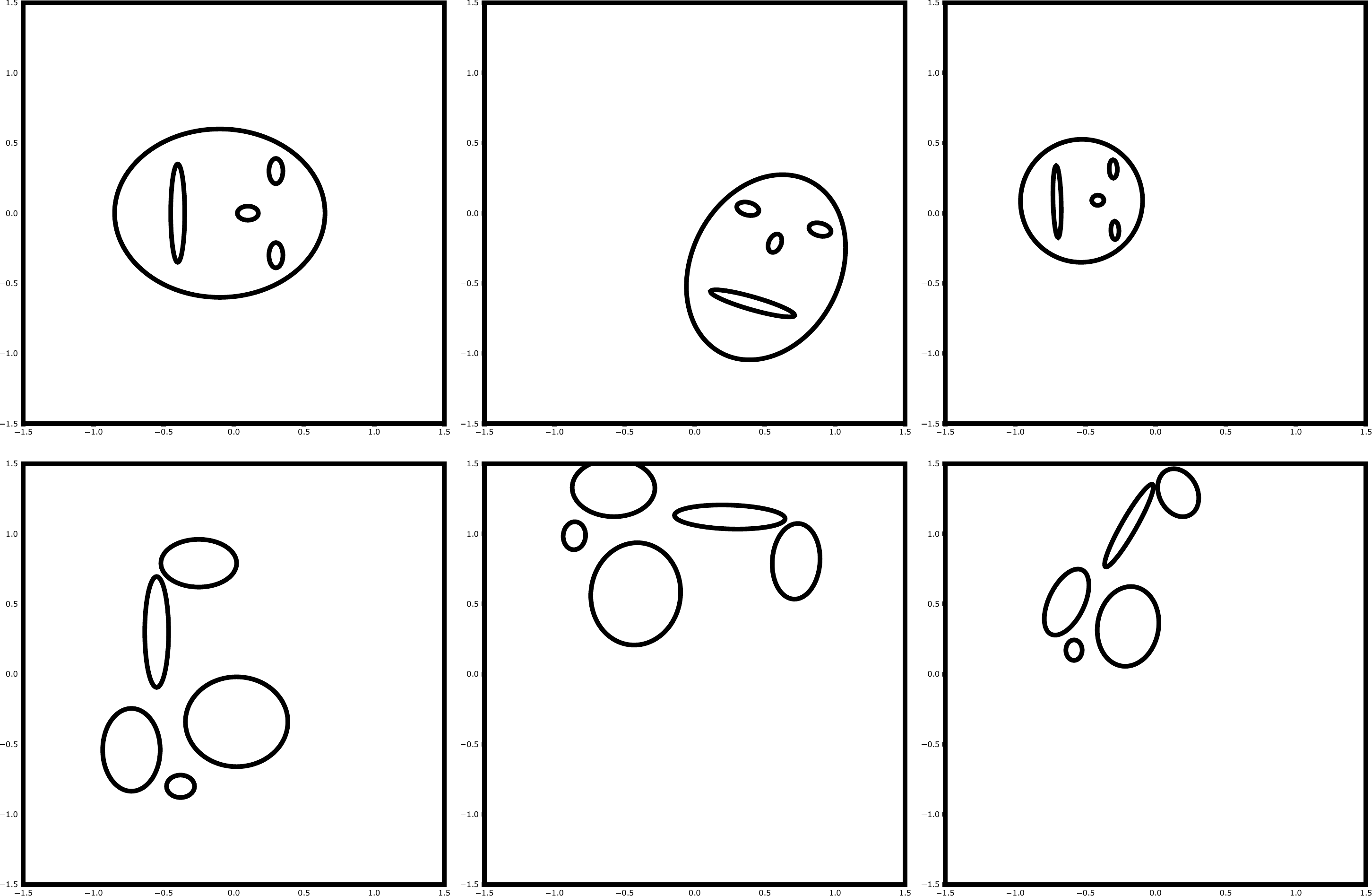}
    \caption{Single Object}
    \label{fig:sample_objects}
    \end{subfigure}
    \begin{subfigure}[b]{0.2\textwidth}
    \centering
    \includegraphics[width=0.8\textwidth]{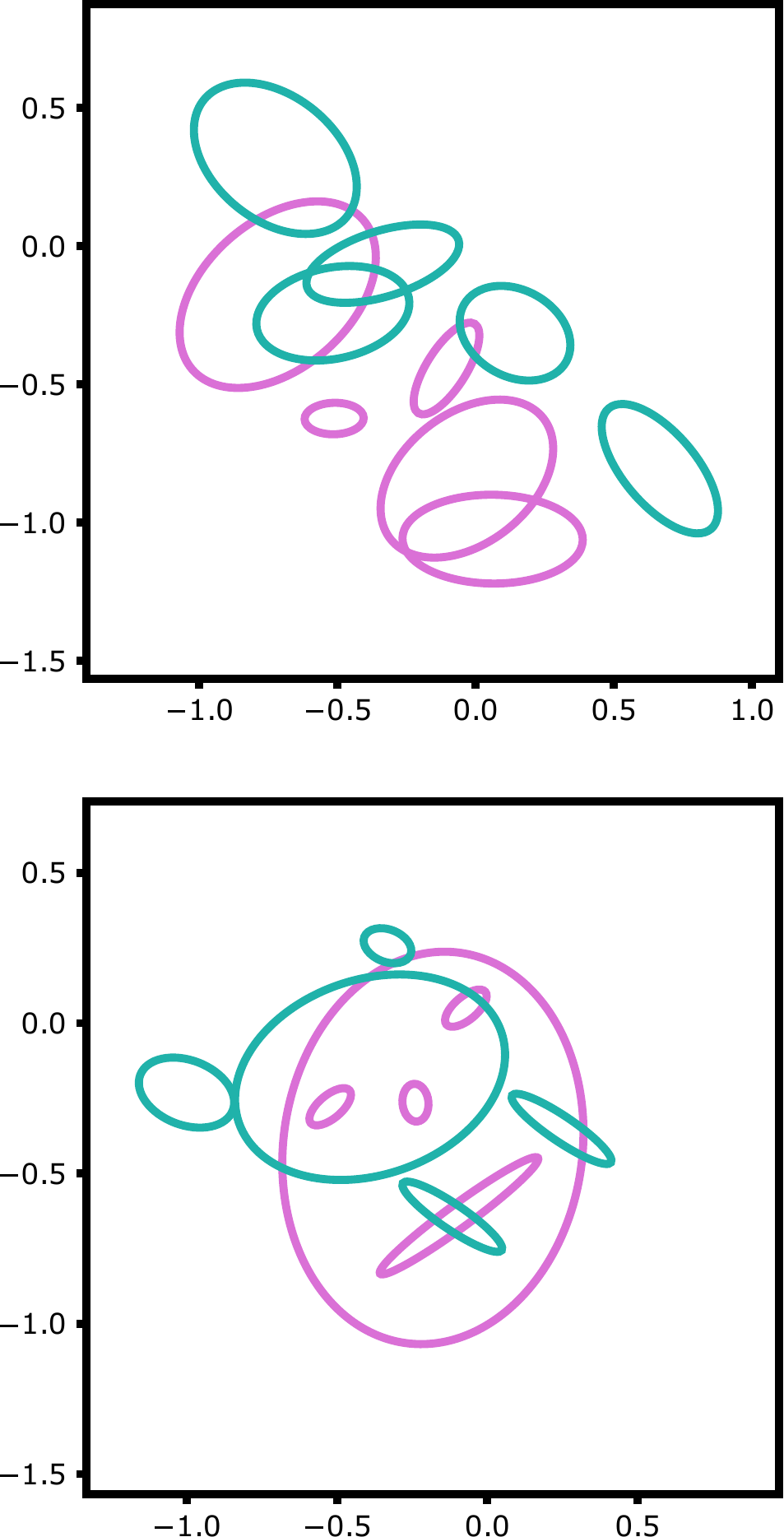}
    \caption{Double Objects}
    \end{subfigure}
    \begin{subfigure}[b]{0.2\textwidth}
    \centering
    \includegraphics[width=0.8\textwidth]{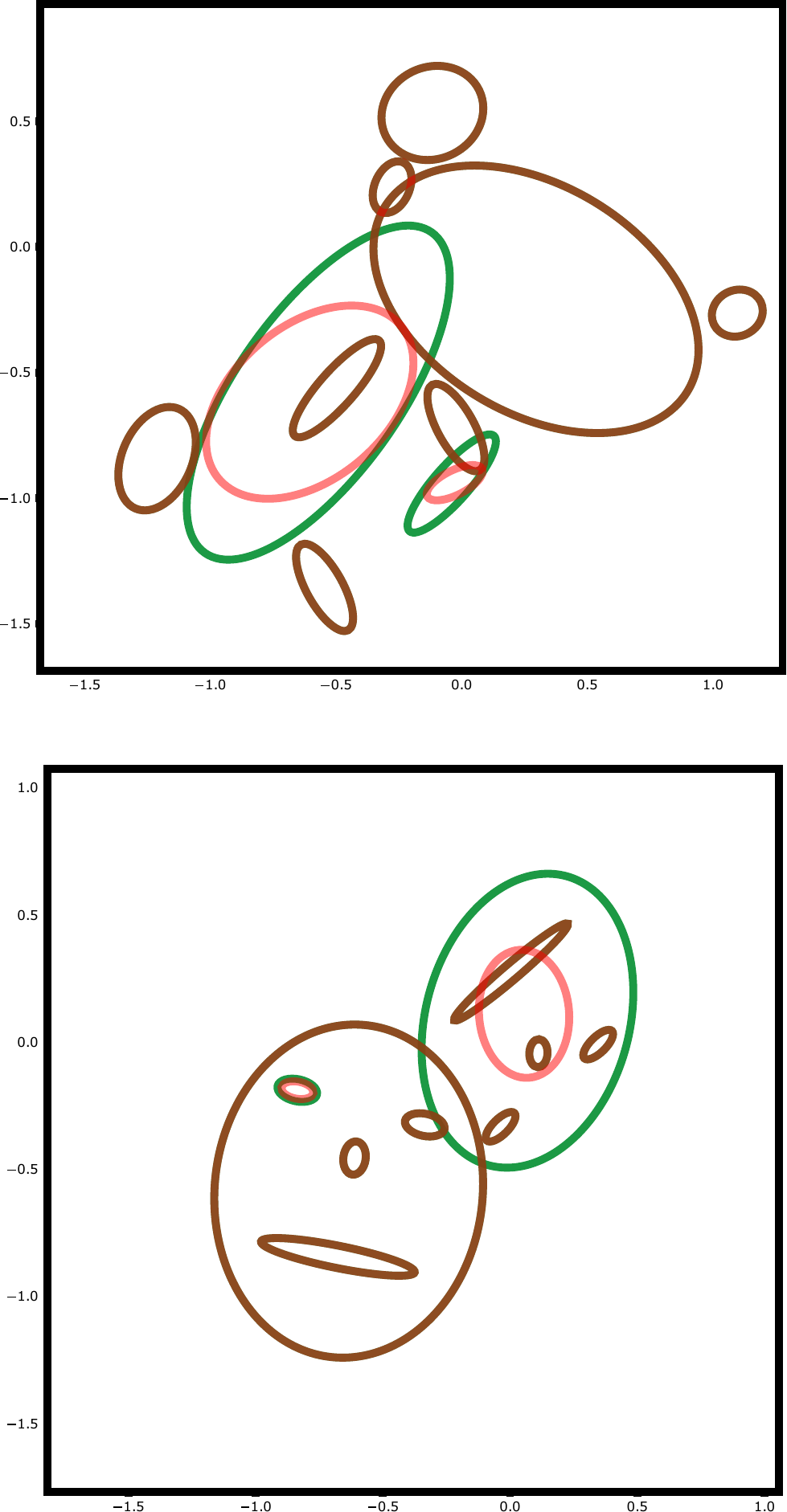}
    \caption{Noisy}
    \label{fig:sample_random_permutations}
    \end{subfigure}
    \caption{Sample "images" in the ellipse world datasets. (\textbf{a}) The first column shows examples of a `face', and a random configuration of 5 ellipses of the type used for an object in the 2-of-20 dataset.  The next two columns demonstrate randomly transformed versions of each "image". (\textbf{c}) Examples of double objects, where the different colours indicate different objects. (\textbf{c}) Examples of the perturbed multi-object experiments. The green ellipses are ground truth and the red ellipses are the input to the model (the brown shows the overlap of the red and green). }
    \label{fig:sample_face_sheep}
\end{figure}

All objects in the ellipse world are comprised of 5 ellipses. The "images" in Figure~\ref{fig:sample_face_sheep} show sample object configurations for face, sheep, etc. Each ellipse is represented by 5 degrees of freedom that specify the affine transformation that converts a unit circle at the origin into the ellipse. (which is calculated from rotation, horizontal scale, vertical scale, horizontal translation and vertical translation). In general, an affine transformation in 2-D has six degrees of freedom and since an ellipse has only 5 degrees of freedom there is a one-dimensuional family of affine transformations that all produce the same ellipse by using different combinations of orientation and shear. In order to prevent ambiguity, all the ellipses are axis aligned when the object is in its canonical pose.   We also have a symbolic representation for the object which encodes spatial position, scale, etc (it is the affine transformation that is applied to the object in its canonical pose to get the object instance in the "image").

The ellipse world consists of a grid where each cell has at most one ellipse center. The x-y spatial domain is divided into non-overlapping cells of size 0.05 and each ellipse is assigned to the closest cell. For example an ellipse with xy center coordinate (0.43, 0.78) will be assigned to (0.45, 0.8) cell. The size of the cells can be varied for different performance. These cells are created in order to allow us to give the model an xy coordinate without giving the model all the information about the xy coordinate of the ellipse.

Skipping pixel level input reduces the computational cost since no computation is needed for columns that do not contain an ellipse.  Also, symbolic part representations are a finer and easier adjustable input medium for quickly evaluating different aspects of a new architecture such as GLOM. Therefore, we propose a new dataset and task for this analysis. 

The training set size is 500,000, with a validation set of size 5000. Each of these examples contain a varying number of objects based on various experiments.
\section{Experiments}
The goal of these experiments is to test whether eGLOM can learn to make highly multi-modal predictions for the whole from highly ambiguous parts, and whether simple attention-weighted averaging at the level of the whole can settle on the shared mode of these predictions. 

This is done by considering several different cases:
\begin{itemize}
    \item {\bf 1-from-2} has two types of possible object (face and sheep) and each "image" only contains one object instance. This case does not require attention because all ellipses belong to the same object.
    \item {\bf 2-from-2} has two types of possible object (face and sheep) and each "image" contains two object instances which may or may not be of different types. Even if they are of the same type, their poses will be different so attention is needed to decide which object-level embeddings at different locations should be averaged.
    \item {\bf 2-from-20} has twenty types of 5-ellipse object generated at random and each "image" contains two object instances.
\end{itemize}

\subsection{Tasks}

\paragraph{Unperturbed "images"} The object instances are generated by applying randomly chosen affine tranformations to the object in its canonical pose.  The object is translated randomly in (-0.75 to 0.75). The rotations are varied randomly 360 degrees, and there is no shear. Scaling is performed randomly on the x and y axes separately. 

The 1-from-2 case where there is only a single object instance per "image" is the easiest, and demonstrates that eGLOM can make highly multi-model predictions for the whole from highly ambiguous parts. This simple case does not need attention enabled as it does not have multiple different objects in the "image". 

The second 2-from-2 case, where there are two different object instances in a single "image", is slightly more difficult, and fails without attention. Once attention is enabled, this case can succeed in separating the two objects, even when highly overlapping, by settling on a the shared mode of the predictions from each set of five ellipses that belong to the same object.


The 2-from-20 case, where there are twenty different object types, is useful as the objects are randomly created, using the same code. Any potential biases from Sheep and Face shapes are removed, and the prediction of the whole from parts and settling on the shared mode of the predictions still succeeds.

\paragraph{Perturbed "images"} A variation of the experiment is to perturb the input ellipses so that they do not exactly match the object. This is done by randomly selecting one or two ellipses within the object, randomly scaling these ellipses, and randomly translating the ellipse centers in such a way that they still stay within the grid cell of the original. They need to stay within the grid cell because the grid cell location is used within the model, and moving the object centre outside of the grid would feed inconsistent data into the model.

This perturbation variation was used to determine whether the model can correct the reconstruction of the perturbed ellipses.

\paragraph{Interpolation} To test the ability of eGLOM to generalize to interpolated transformations, we divide the $[0^{\circ},360^{\circ}]$ range into four $90^{\circ}$ segments. Two non-adjacent segments are used for sampling training rotations and the other two for uniformly sampling the test rotations.

\subsection{An Autoencoder Baseline} \label{sec:baseline}
\begin{wrapfigure}{r}{0.3\textwidth}
    \centering
    \includegraphics[width=0.3\textwidth]{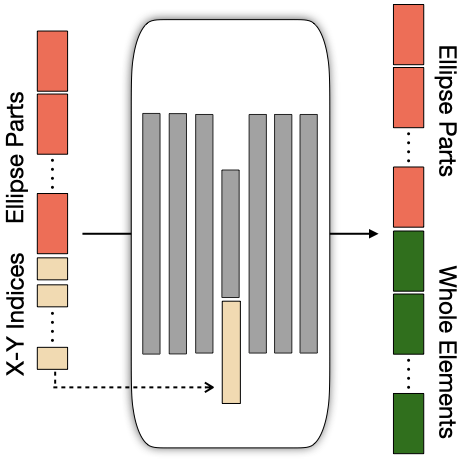}
    \caption{Baseline auto encoder architecture.}
    \label{fig:baseline}
\end{wrapfigure}

As a basic baseline for our studies we consider a multi-layer autoencoder. The input and desired output are made by concatenating the 6-dimensional symbolic representations of the objects and parts in our datasets. All the layers in baseline are 1024-D fully connected and using the ReLU non-linearity with one 512D encoding layer. The final layer predicts the whole parameters in addition to reconstructing the input ellipses. Fig.~\ref{fig:baseline} visualizes the connections between input, hidden layers, and the output dimensions.

Unlike GLOM, an MLP autoencoder is not permutation invariant with respect to the input ellipses. Therefore, we make the tasks easier for the autoencoder baseline by always providing a fixed ordering of parts relative to the wholes. For example, the nose is always the first ellipse of a face ellipse group, so if the first object is a face, the first ellipse is always the nose and if the second object is a face, the 6th ellipse is always the nose.


Furthermore, since in GLOM the grid cell indices are used for predicting the ellipses from wholes, we provide the grid cell indices of ellipses in addition to their exact coordinates as the input. We also concatenate the flattened grid indices to the middle layer encoding before decoding the concatenated embedding. 
 \begin{figure}[t]
    \centering
    \includegraphics[width=\textwidth]{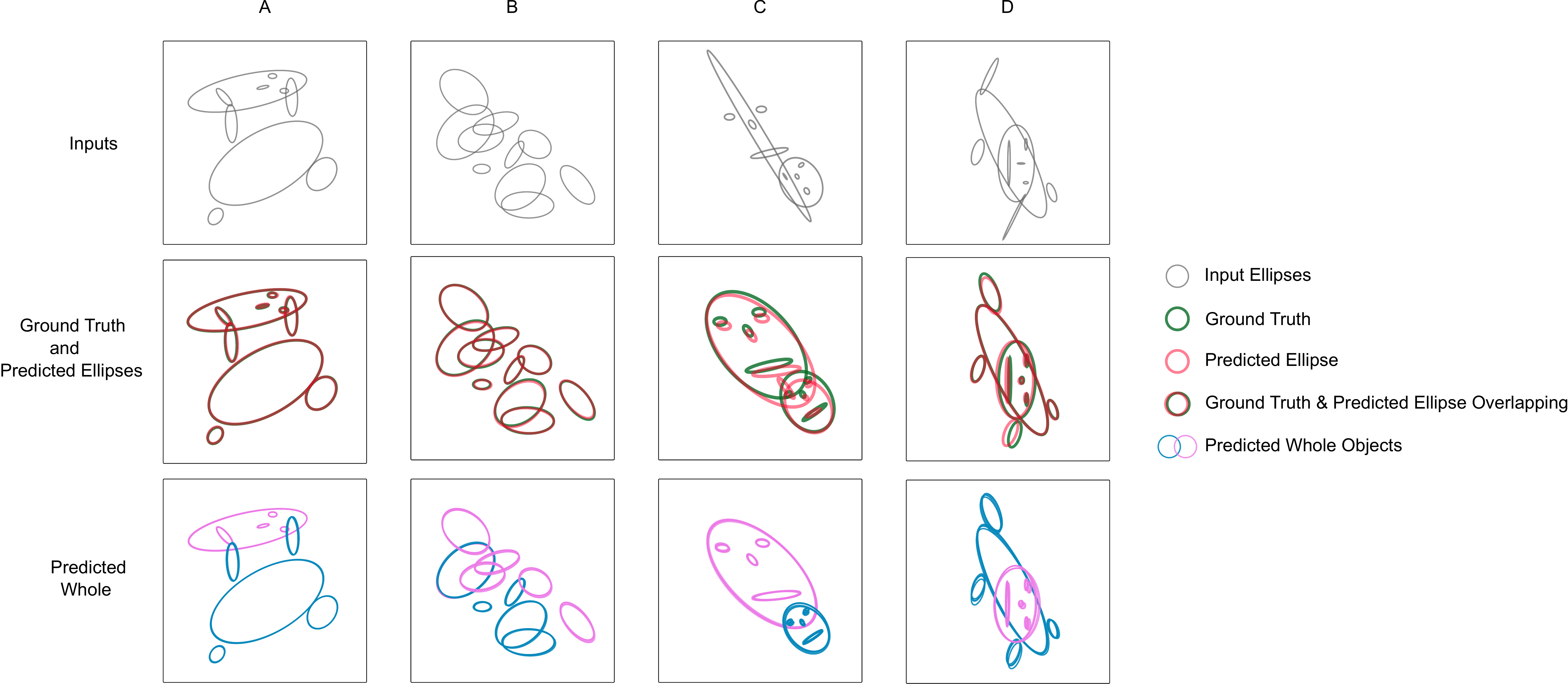}
    \caption{Examples of the results of the three experiments.
    A) 2-from-2: The randomly chosen objects happen to be a face and a sheep. eGLOM reconstructs the whole nearly exactly.
    B) 2-from-20: Two randomly chosen objects out of the twenty randomly created ones. eGLOM is able to reconstruct, and separate, the objects nearly perfectly.
    C and D) Perturbations of the 2-from-20 experiment: The perturbed ellipses are given as input. The ellipses are nearly predicted correctly, and although there is more disagreement about the whole than in 2-from-2 and 2-from-20, there is still general agreement.  }
    \label{fig:example_outputs}
\end{figure}

\subsection{Results}

In order to test eGLOM's ability to disentangle and associate parts based on their part-whole relationships, we first start with 2-from-2 and 2-from-20 datasets where 2 objects (10 ellipses) are present in each "image". Then we evaluate eGLOM's robustness to random perturbations of the parts and out-of-distribution poses of the wholes. 

\paragraph{Unperturbed "images"} Figure ~\ref{fig:example_outputs} shows that given raw ellipses, eGLOM is able to reconstruct ellipses, predict correct part-whole (ellipse-object) assignments, and predict the correct object-level parameters. The last row shows the visualized wholes based on the predicted whole parameter for each ellipse. It verifies that all parts are able to predict correct whole parameters and whole classification individually for both 2-from-2 and 2-from-20 tasks. 

The quantitative results (Mean Squared Error) compared against the autoencoder baseline described in \ref{sec:baseline} are reported in Table~\ref{tab:glom_vs_baseline}. In both cases, eGLOM outperforms the baseline by 2 orders of magnitude. Since our baseline was significantly under-performing for 2-from-20, we base our comparison on 1-from-20 (where only one object out of 20 types are present in each "image"). A parameter sweep was performed for both 1-from-20 and 2-from-2. The only difference shown was that in 2-from-2, constricting the embedding size half way through the network led to better Whole MSE scores (though it hurt the Part MSE). We report the numbers that balance the two MSE. In 1-from-20, further constriction of the embedding size half way through led to degradation of the model overall. 
\begin{table}[t]
\centering
\caption{MLP autoencoder baseline vs eGLOM for tasks 2-from-2 and 1-from-20. First, 2-from-2: two random objects, each either a face or a sheep, containing five ellipses. Second, 1-from-20: a single random shape, selected from a twenty randomly generated shapes, each containing five ellipses. The ellipses for each shape are ordered such that the ellipses are always given in the same ordering (ie. first the two eyes, then the mouth, etc).}
\label{tab:glom_vs_baseline}
\begin{tabular}{@{}lllllll@{}}
\toprule
         & \multicolumn{3}{c}{2-from-2}                            & \multicolumn{3}{c}{1-from-20}       \\ \midrule
         & Whole MSE  & Part MSE   & \multicolumn{1}{l|}{Params}  & Whole MSE  & Part MSE   & Params  \\
Baseline & $6.1 e^{-4}$ & $6.1 e^{-4}$ & \multicolumn{1}{l|}{$5.4 e^6$} & $1.0 e^{-2}$ & $4.9 e^{-3}$ & $5.4 e^6$ \\
eGLOM     & $9.6 e^{-6}$ & $7.5 e^{-6}$ & \multicolumn{1}{l|}{$2.3 e^6$} & $2.6 e^{-5}$ & $1.2 e^{-5}$ & $2.0 e^6$ \\ \bottomrule
\end{tabular}
\end{table}

\paragraph{Perturbed "images"} In this task one or two randomly selected input ellipses are perturbed, thus not exactly matching the object. The 2-from-2 data setup is used to evaluate this, and as it is much harder than unperturbed "images" the reconstructions of the whole are not perfect. Figure ~\ref{fig:example_outputs} shows two examples (columns C and D) that are among the most extreme perturbations. Islands are still formed when predicting the whole, the whole classification is correct, and the predicted ellipses are close to the ground truth, thus correcting the perturbed input. 

The MLP autoencoder's performance on this task was significantly worse. Therefore as a basis of comparison, we consider the baseline's performance on 1-from-20, which is an easier task compared to the perturbed task. The Whole MSE of eGLOM in the perturbed task is $9.9 e^{-5}$ and the Part MSE is $3.5 e^{-5}$. For both Part and Whole MSE eGLOM has an error two orders of magnitude lower than the autoencoder baseline on 1-from-20, while tackling a much harder task.

These results indicate that eGLOM learns part-whole relationships successfully and is robust to noise thanks to relying on clustering and the agreement of multiple parts. eGLOM is able to correct noise in one of the parts by first forming a clustering for the correct whole from the other parts. Then it utilizes the learned whole-part relationship to correct the noise in the corrupted part. So, it not only predicts the correct whole, but it can reconstruct what the parts should have been too.

\begin{wrapfigure}{r}{0.35\textwidth}
    \vspace{-15pt}
    \centering
      \centering
      \includegraphics[width=0.35\textwidth]{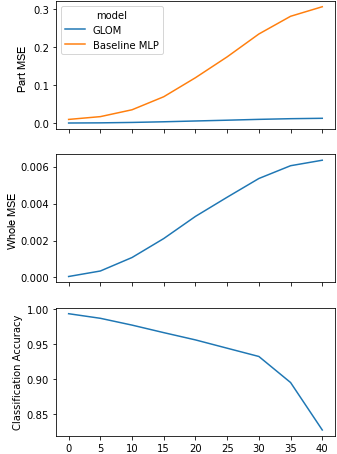}
      \caption{Performance of eGLOM and the MLP Autoencoder baseline as the distance from seen training rotation degrees increase to $45^{\circ}$.}
      \label{fig:interpolation_variations}
    \vspace{-25pt}
\end{wrapfigure}
\paragraph{Interpolation}
In this task we evaluate eGLOM's ability to generalize to unseen viewpoints during test. One of the major benefits in learning part-whole relationships is the ability to generalize to new viewpoints and transformations by using viewpoint equivariant representations of parts and wholes. 

Figure ~\ref{fig:interpolation_variations} shows eGLOM's performance as the test datapoints get farther from the training distribution. Since the $360^{\circ}$ rotation range is divided into $90^{\circ}$ ranges and non-adjacent ranges are used for training, the farthest a test point can be from any training point is $45^{\circ}$. 

In the interpolation task the MLP Autoencoder baseline was able to reconstruct the ellipses to some degree, but failed to classify or predict the whole parameters. Therefore, we only visualize the comparison between eGLOM interpolatibilty and the baseline for the Part MSE (first row). Figure~\ref{fig:interpolation_variations}'s first row shows that eGLOM is exponentially better than baseline in handling new viewpoints. For whole prediction, the second and third rows show that although eGLOM suffers when faced with unseen rotation but it is still able to classify and predict the whole to a certain degree.

\subsection{Ablations}
eGLOM consists of several orthogonal contributions. In this section we explore the effects of each of its architectural and methodological parts. Further ablation studies are discussed in the supplementary materials.

\begin{figure}
\centering
    \begin{subfigure}[t]{0.4\textwidth}
    \centering
    \includegraphics[width=1\textwidth]{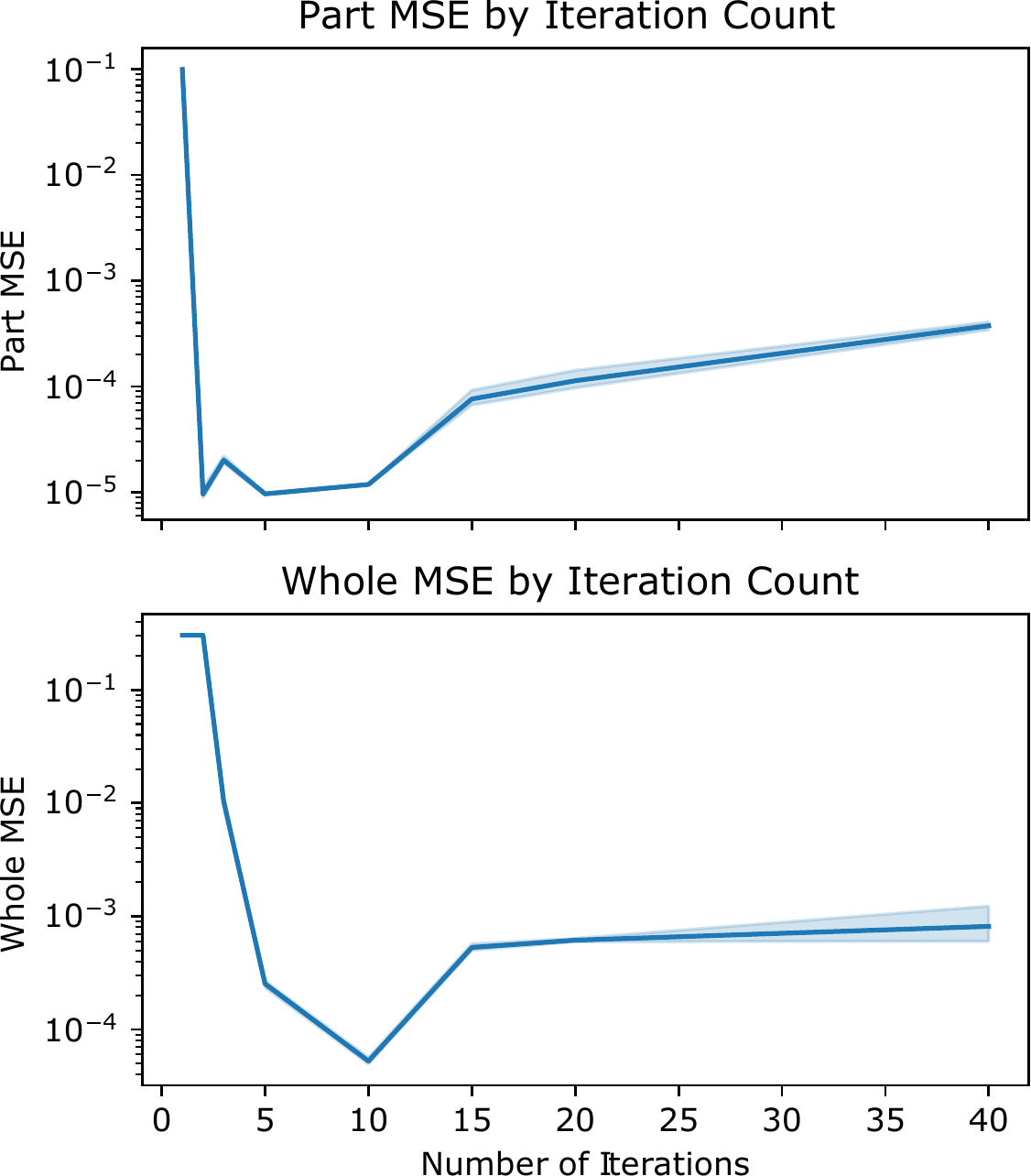}
    \caption{Number of iterations.}
    \label{fig:iteration_variations}
    \end{subfigure}
    \hspace{+10pt}
    \begin{subfigure}[t]{0.4\textwidth}
    \centering
    \includegraphics[width=1\textwidth]{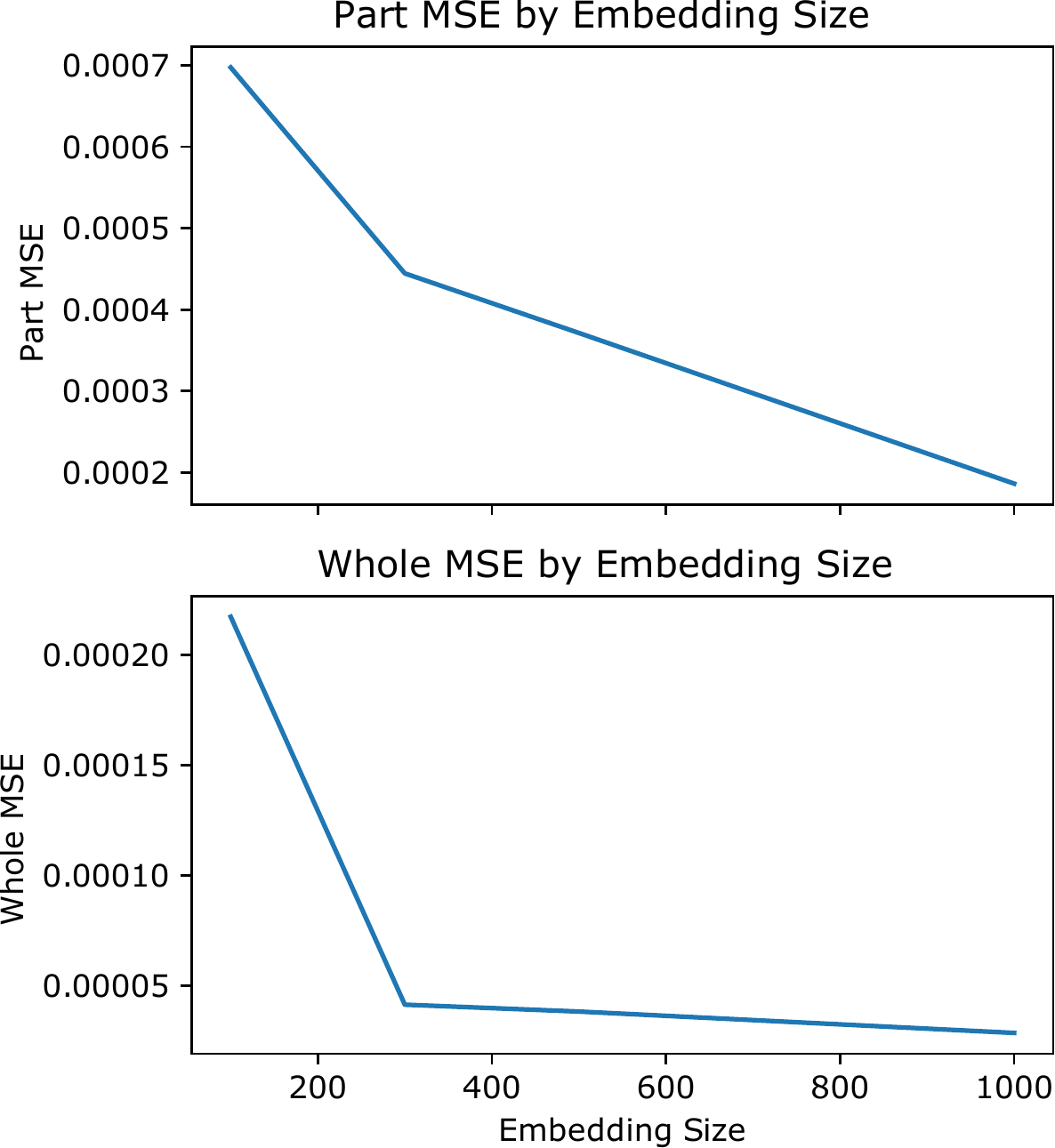}
    \caption{Embedding size.}
    \label{fig:embedding_size_variations}
    \end{subfigure}
    \caption{Part and Whole MSE of eGLOM on the 2-from-20 task for various number of iteration and embedding dimensions.}
\end{figure}

\paragraph{Iteration Count} 
One major component of eGLOM is the iterative aspect of updating the part and whole embeddings based on the top-down and bottom-up predictions and the attentional contributions from the same level. In our experiments we vary the number of iterations up to 40. The results are visualized in Figure~\ref{fig:iteration_variations}.

We observe that eGLOM requires at least 5 iterations to converge. Moreover, extra iterations can increase the MSE. We hypothesize that the increase in MSE is due to drifting of the input ellipse symbols away from their original correct values. 
\footnote{The ellipse symbols are not clamped at their original values throughout the iterations because we want to allow perturbed ellipses to be corrected.} The classification accuracy does not suffer from an increased number of iterations in these two tasks.

\paragraph{Embedding Dimension}
Another aspect is the required embedding dimension for eGLOM to disentangle the multi-modal part-whole distributions. Figure~\ref{fig:embedding_size_variations} shows that as the embedding size increases the task becomes easier. The lowest embedding dimension for which the model converges is 100.

\begin{figure}
\centering
 \begin{subfigure}[t]{0.4\textwidth}
    \centering
    \includegraphics[width=1\textwidth]{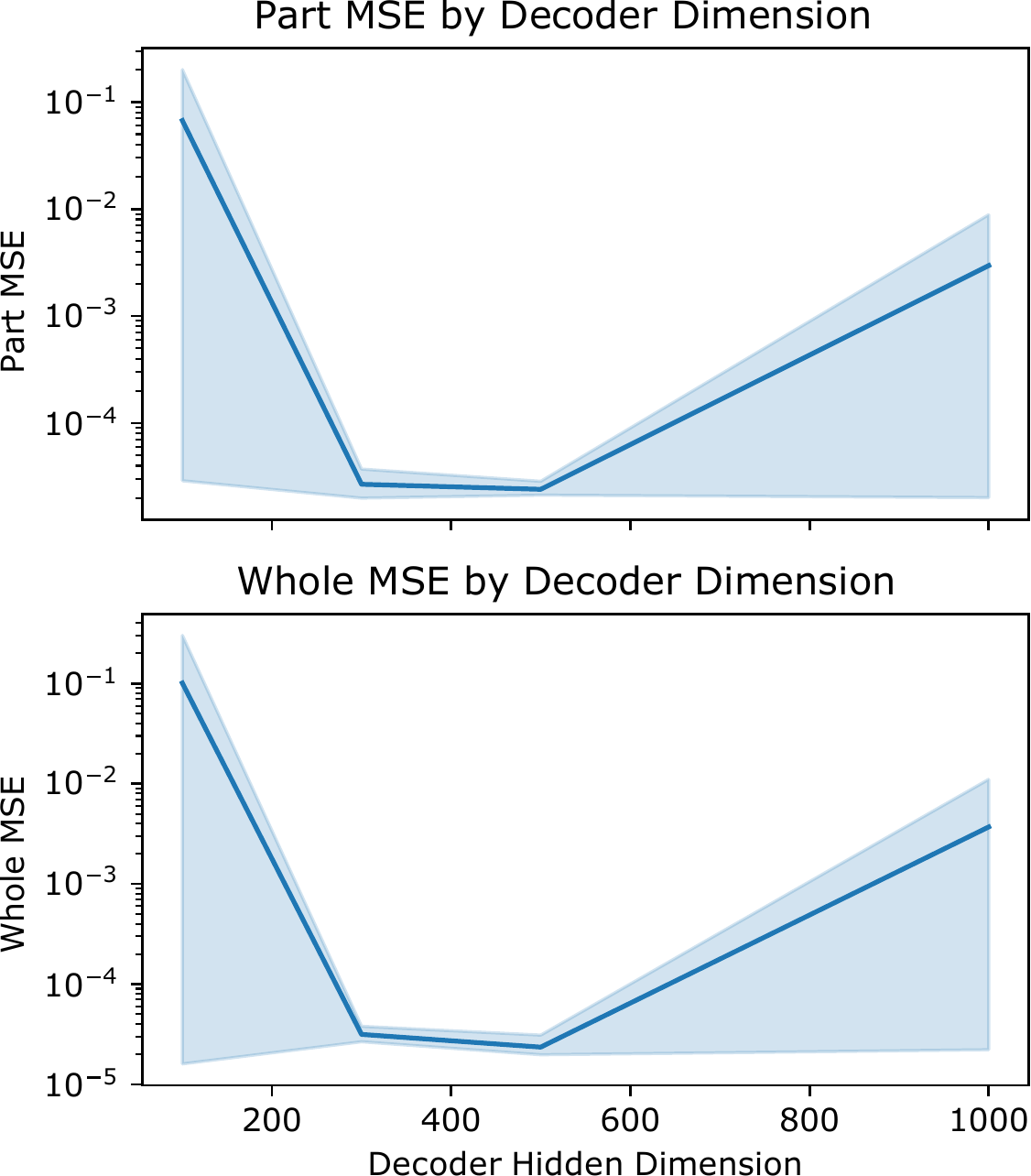}
    \caption{Dimension of the top-down decoder.}
    \label{fig:decoder_dimension_variations}
    \end{subfigure}
    \begin{subfigure}[t]{0.55\textwidth}
    \centering
    \includegraphics[width=1\textwidth]{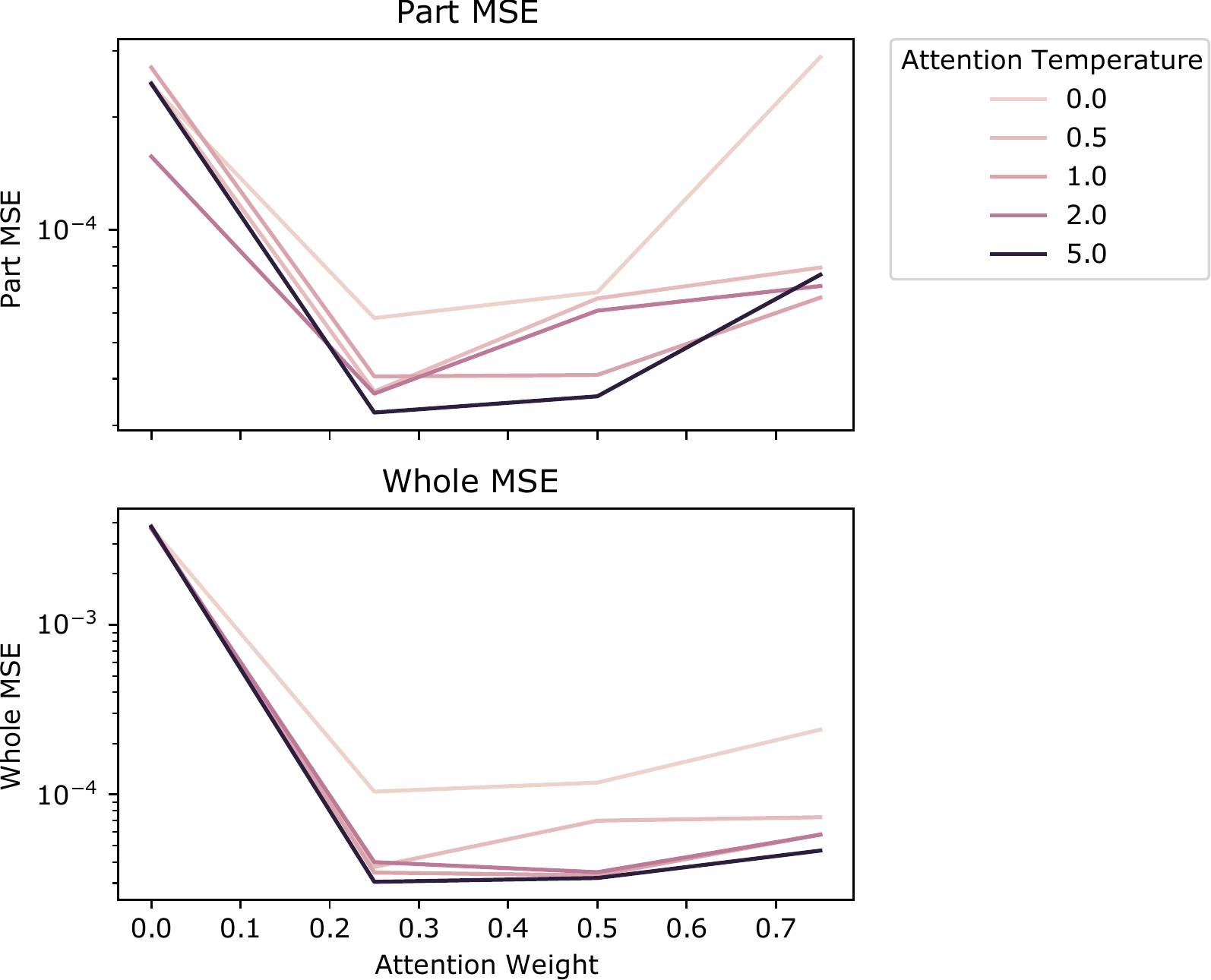}
    \caption{Attention weight and temperature}
    \label{fig:attention_variations}
    \end{subfigure}
    \caption{Part and Whole MSE of eGLOM on the 2-from-2 task for various decoder dimensions and attention weights and temperature.}
    
\end{figure}

\paragraph{Decoder Dimensions} In our experiments we found that unlike embedding size, the top-down decoder hidden dimension has an optimal size around 2x the embedding size. Figure~\ref{fig:decoder_dimension_variations} visualizes how Part and Whole MSE changes based on the decoder dimension.

\paragraph{Attention} The implicit clustering in eGLOM is implemented by the attention function. Therefore, incorporating an attention temperature can change the clustering assignments from soft to hard. In practice, we observe that a higher temperature results in better MSE as visualized by Figure~\ref{fig:attention_variations}. In this study we also vary the weight of the attentional contribution relative to the bottom-up and history contributions. Figure~\ref{fig:attention_variations} indicates that attention is a crucial component of the model as increasing the attention weight to at least $0.3$ is necessary for a low MSE.

\section{Conclusion and Discussion}
We propose a series of tasks and experiments to analyze and design specifics of a GLOM architecture.
The goal of these experiments is to test whether GLOM can learn to make highly multi-modal predictions for the whole from highly ambiguous parts, and whether simple attention-weighted averaging at the level of the whole can settle on the shared mode of these predictions. 

This is done by considering several different cases: two object types and one object per input, where all the ellipses are a part of the whole; two object types and two objects per input, where half the ellipses are part of one and the other half part of the other; and twenty randomly created object types and two objects per input. We generate different variations of our dataset to experiment on each case.

We compare eGLOM against an MLP Autoencoder model as the baseline. The results indicate eGLOM is significantly better in predicting the whole parameters from scattered and entangled parts. Furthermore, we show that eGLOM does perform well in even harder tasks such as generalization to unseen viewpoints. Our results indicate eGLOM is able to make highly multi-modal predictions for the whole from highly ambiguous parts. 

Further methodological improvements and stronger baselines would strengthen our observations significantly. First, our setup is currently fully supervised and there is a strong potential for a self-supervised setup, such as incorporating a contrastive regularizer. Once eGLOM is able to be trained in a self-supervised fashion, moving to pixel based input is the next step. This would increase the dimensionality of the input from five or ten 6D vectors to a 3D vector for every pixel in the image. 

Furthermore, implementing a transformer based baseline to compare against would better showcase GLOM's pros/cons, and the difference in performance between a transformer and eGLOM could provide further insight into the workings of GLOM.  

In summary, our study shows that GLOM can learn to make highly multi-modal predictions for the whole from highly ambiguous parts, while also being able to reconstruct the parts from the whole. Which strongly motivates further studies on self-supervision, and on how to adopt complex images as the input.
\bibliographystyle{plainnat}
\bibliography{ref}

\newpage
\section{Supplementary}
\subsection{Full architecture}

\paragraph{Weight Variations}

\begin{figure}
    \begin{subfigure}[b]{0.47\textwidth}
    \centering
    \includegraphics[width=1\textwidth]{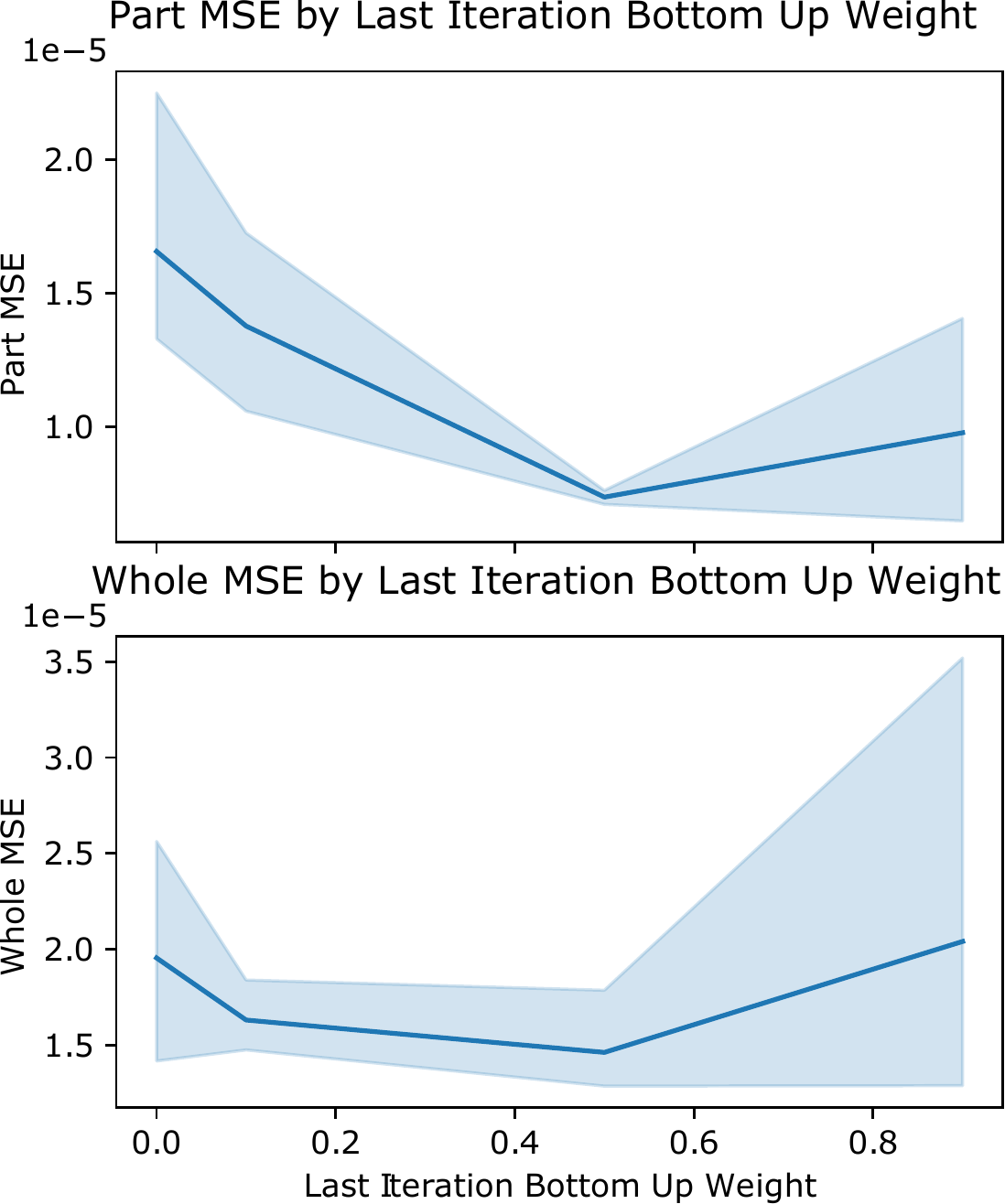}
    \caption{2-from-2}
    \end{subfigure}
    \begin{subfigure}[b]{0.5\textwidth}
    \centering
    \includegraphics[width=1\textwidth]{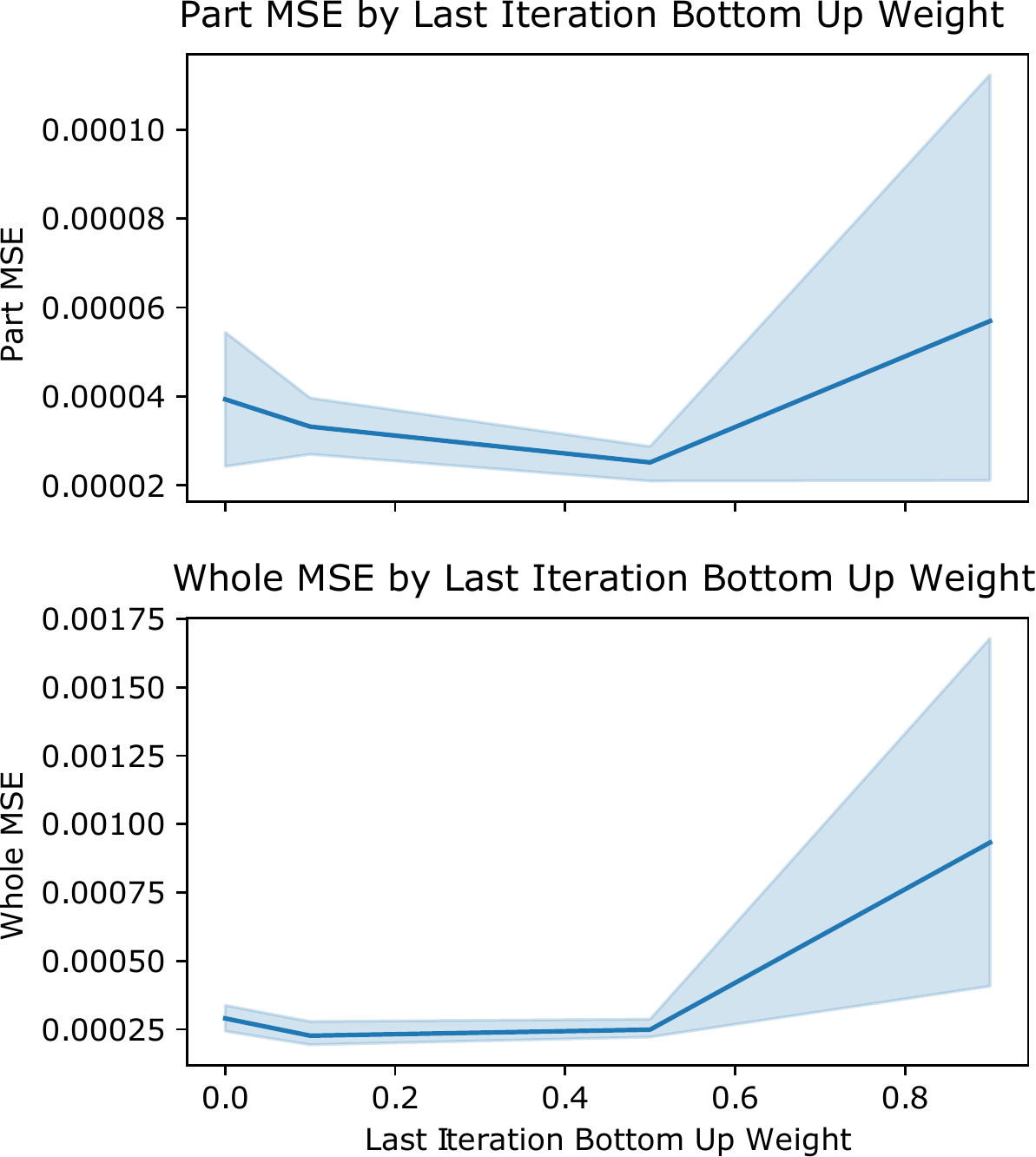}
    \caption{2-from-20}
    \end{subfigure}
    \caption{Ablation of the amount of weight that is given to the bottom-up connection (from the ellipse symbol layer to the ellipse embedding layer vs the top-down connection from the whole embedding layer, or from the ellipse embedding layer to the whole embedding layer vs the attention) on the last iteration. Of course, in the cases where the bottom-up weight is non-zero, a full top-down connection is run in order to generate the final ellipse symbols that are reported. The Classification Accuracy is not shown, as it is mostly noise. Multiple models are trained and the 90\% confidence intervals are shown.}
    \label{fig:end_bu_weight_variations}
\end{figure}

\begin{figure}
    \begin{subfigure}[b]{0.5\textwidth}
    \centering
    \includegraphics[width=1\textwidth]{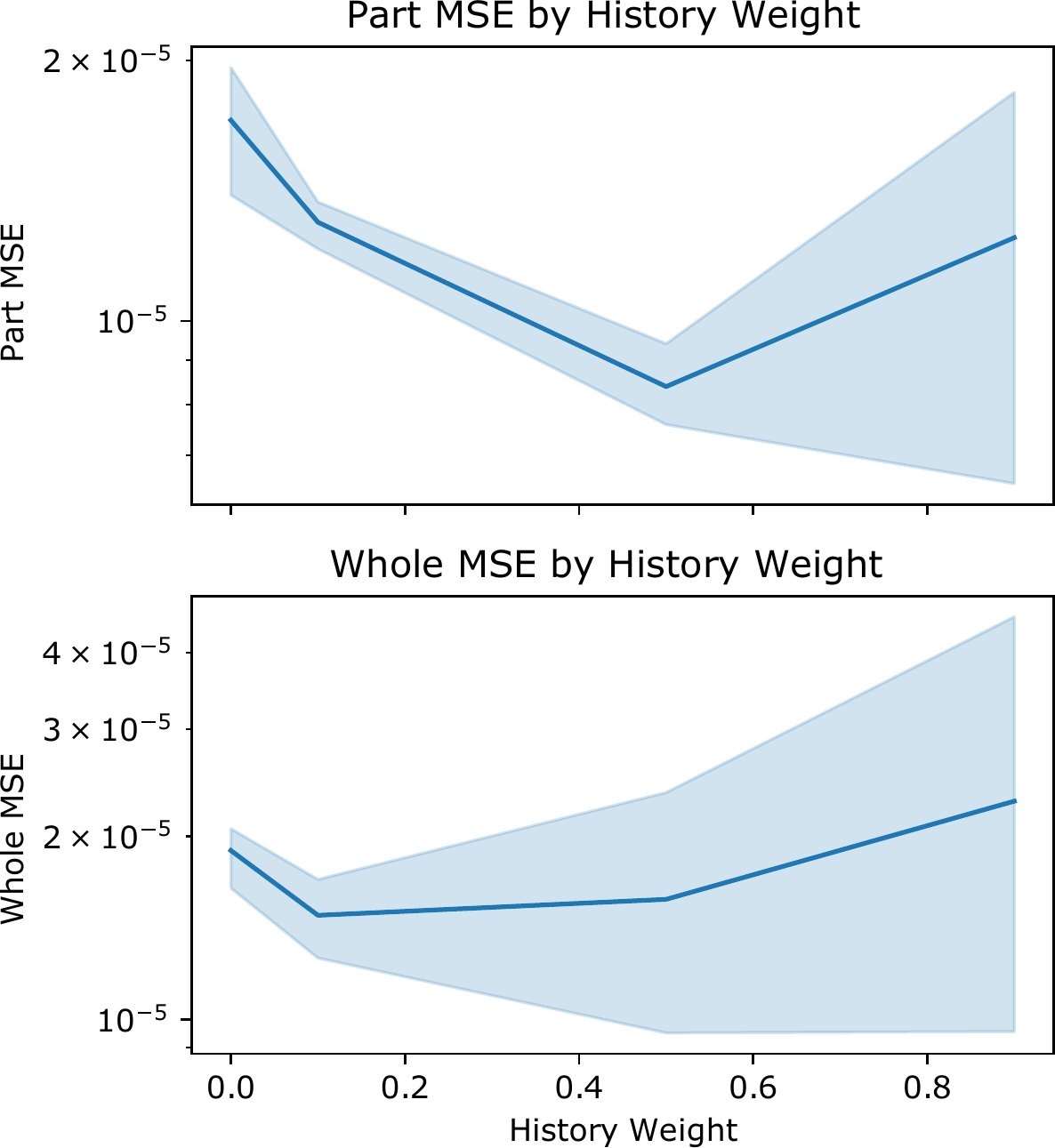}
    \caption{2-from-2}
    \end{subfigure}
    \begin{subfigure}[b]{0.5\textwidth}
    \centering
    \includegraphics[width=1\textwidth]{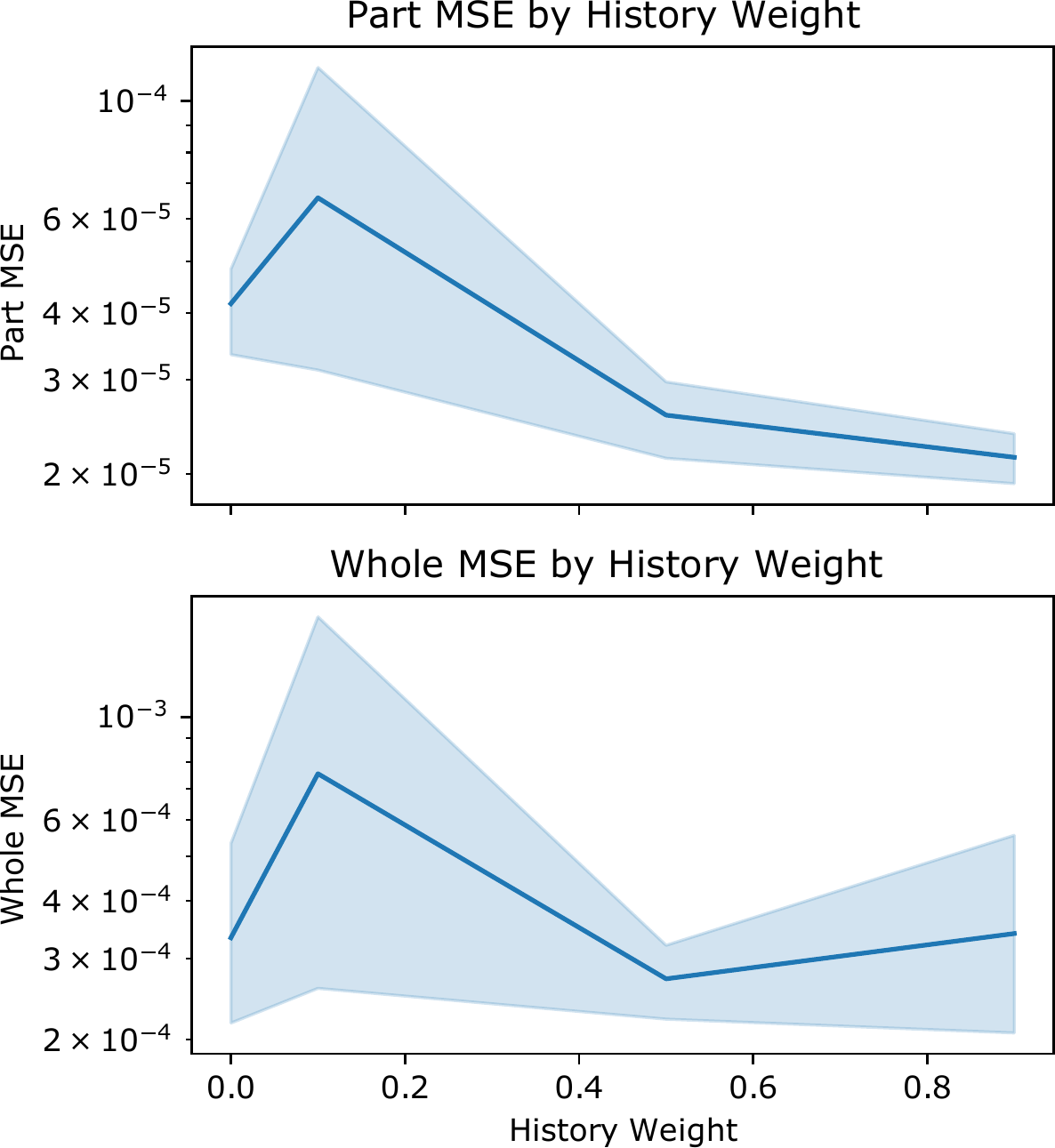}
    \caption{2-from-20}
    \end{subfigure}
    \caption{Ablation of Attention Weight and Temperature on the 2-Basic and 2-Large datasets. The Classification Accuracy is not shown, as it is mostly noise. Multiple runs are completed, and the 90\% confidence intervals are shown.}
    \label{fig:history_weight_variations}
\end{figure}

Figure ~\ref{fig:end_bu_weight_variations} and Figure ~\ref{fig:history_weight_variations} show the affects of the bottom-up contribution and history contribution on the end results.

\paragraph {Selected Parameters}

Table ~\ref{tab:glom_vs_baseline_parameters} shows the parameters used for the numbers reported in Table ~\ref{tab:glom_vs_baseline}.

\begin{table}[t]
\centering
\caption{Parameters used for the Autoencoder baseline and eGLOM.}
\label{tab:glom_vs_baseline_parameters}
\begin{tabular}{@{}llllll@{}}
\toprule
          \multicolumn{3}{c}{Baseline}                            & \multicolumn{3}{c}{eGLOM}       \\ \midrule
         & Bottleneck Size  & \multicolumn{1}{l|}{Layer Size}  & Iterations & Embedding Size & Decoder Dimension   \\
2-from-2 & 512 & \multicolumn{1}{l|}{1024} & 10 & 500 & 500 \\
1-from-20     & 512 & \multicolumn{1}{l|}{1024} & 10 & 500 & 300\\ \bottomrule
\end{tabular}
\end{table}

\subsection{Creation of Islands}
\paragraph{TSNE}

In order to show how the embedding layers change over time, we collect the embedding vectors at each iteration, and use TSNE to embed them into a 2D plane. We do this for the 2-from-2 experiment, and use the hyper-parameters reported in Table ~\ref{tab:glom_vs_baseline_parameters}. 

\begin{figure}
    \centering
    \includegraphics[width=1\textwidth]{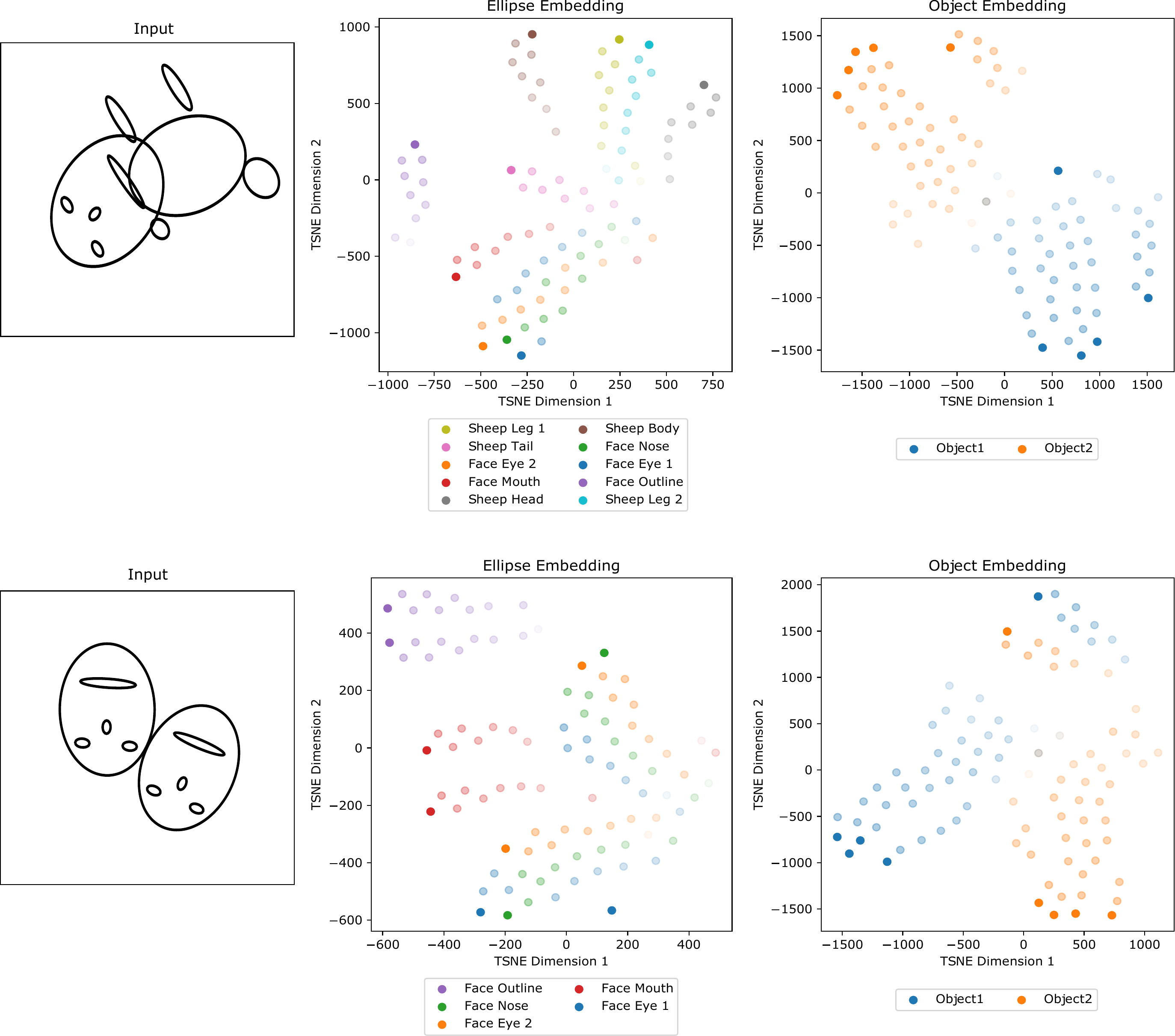}
    \caption{Two examples of how 2-from-2 ellipse and object embeddings change over the 10 iterations. Each dot indicates the TSNE embedding of a single positional embedding at a single iteration. The opacity of the dot shows the iteration - with later iterations being more opaque.}
    \label{fig:ellipse_obj_embeddings}
\end{figure}

Figure ~\ref{fig:ellipse_obj_embeddings} shows two example inputs, and how eGLOM forms islands at each level. It can be seen that ellipse embeddings can be very close together if they are similar (eg. mouths can be close together), but over time at the object embedding level the objects are separated and the different objects form islands.

 
\end{document}